# LATTA: Langevin-Anchored Test-Time Adaptation for Enhanced Robustness and Stability

Harshil Vejendla
Department of Computer Science
Rutgers University
New Brunswick, NJ, USA
harshil.vejendla@rutgers.edu

*Abstract*—Test-time adaptation (TTA) aims to adapt a pre-trained model to distribution shifts using only unlabeled test data. While promising, existing methods like Tent suffer from instability and can catastrophically forget the source knowledge, especially with small batch sizes or challenging corruptions. We argue that this arises from overly deterministic updates on a complex loss surface. In this paper, we introduce Langevin-Anchored Test-Time Adaptation (LATTA), a novel approach that regularizes adaptation through two key mechanisms: (1) a noisy weight perturbation inspired by Stochastic Gradient Langevin Dynamics (SGLD) to explore the local parameter space and escape poor local minima, and (2) a stable weight anchor that prevents the model from diverging from its robust source pre-training. This combination allows LATTA to adapt effectively without sacrificing stability. Unlike prior Bayesian TTA methods, LATTA requires no architectural changes or expensive Monte-Carlo passes. We conduct extensive experiments on standard benchmarks, including Rotated-MNIST and the more challenging CIFAR-10-C. Our results demonstrate that LATTA significantly outperforms existing methods, including Tent, CoTTA, and EATA, setting a new state-of-the-art for self-supervised TTA by improving average accuracy on CIFAR-10-C by over 2% while simultaneously reducing performance variance.

## I. INTRODUCTION

Neural networks deployed in the wild frequently encounter data from distributions different from their training set, leading to a significant drop in performance. Test-time adaptation (TTA) has emerged as a powerful paradigm to counter this by adapting the model using only the streaming, unlabeled test data. Methods like Tent [1] achieve this by minimizing the entropy of model predictions, effectively increasing the model's confidence on the target domain.

Despite their promise, such self-supervised TTA methods face critical challenges. They are often brittle, sensitive to batch size, and prone to *catastrophic forgetting*, where the model over-fits to a few target batches and diverges irreversibly from its capable initial state [2]. While more recent methods like CoTTA [2] or EATA [7] introduce mechanisms like weight averaging or explicit regularization to combat this, they can still be susceptible to noisy gradients from challenging shifts or small batches.

We hypothesize that a core issue is the deterministic nature of the gradient updates, which can easily get trapped in poor local minima of the self-supervised loss landscape. To address this, we introduce **Langevin-Anchored Test-Time Adaptation (LATTA)**. Our method reimagines the TTA update step as a single-step Bayesian posterior sampling. Specifically, we make two crucial contributions:

1) **Langevin Weight Perturbations:** After each entropy-minimizing gradient step, we inject carefully scaled Gaussian noise into the model weights. This realizes one step of Stochastic Gradient Langevin Dynamics (SGLD) [4], encouraging the model to explore the local parameter space and avoid sharp, unstable minima.
2) **Stable Weight Anchor:** To ground this exploration and prevent catastrophic forgetting, we maintain an exponential moving average (EMA) of the adapted weights, which serves as a stable anchor. The model is gently pulled towards this anchor during adaptation, blending exploratory updates with robust source knowledge.

The combination of exploration and anchoring makes LATTA both effective and stable. It requires no changes to the model architecture and adds negligible computational overhead. We perform a thorough evaluation on the Rotated-MNIST and CIFAR-10-C benchmarks. LATTA not only achieves superior accuracy compared to strong baselines like Tent, CoTTA, and EATA, but also demonstrates lower variance across different corruptions and data orderings. Our work establishes a new state-of-the-art for self-supervised TTA and highlights the power of combining stochastic weight-space exploration with stable anchoring.

## II. RELATED WORK

**Self-Supervised TTA.** Test-time adaptation using self-supervision was popularized by Tent [1], which adapts batch normalization parameters by minimizing prediction entropy. TENT is simple and effective but can be unstable. Subsequent work has focused on improving robustness. CoTTA [2] mitigates forgetting by using a stochastic weight average and resetting the model to a teacher model when errors accumulate. EATA [7] introduces an elastic consolidation regularizer that penalizes changes to parameters deemed important for the source task, alongside a sample-rejection mechanism based on entropy. Our method, LATTA, also aims for stability, but achieves it through a novel combination of stochastic weight exploration and a continuous EMA anchor, rather than resets or parameter-specific penalties.

**Bayesian TTA.** A natural way to handle uncertainty at test-time is through a Bayesian lens. B-TTA [3] proposes adding variational heads to the network and performing Monte-Carlo sampling at inference time. While principled, this approach adds significant architectural complexity and computational cost. LATTA offers a lightweight alternative, achieving the regularizing effect of a Bayesian posterior sample via a single SGLD step without any new modules.

**Langevin Dynamics in Deep Learning.** Stochastic Gradient Langevin Dynamics (SGLD) [4] is a seminal algorithm for sampling from a posterior distribution. It has been used primarily for Bayesian training of neural networks. More recently, "Langevin Smoothing" [6] applied SGLD during the *training phase* as a defense against adversarial attacks, using the noise to smooth the loss landscape. **Our work is fundamentally different**: we apply SGLD-style updates at *test time* for adaptation to natural distribution shifts, not at training time for adversarial robustness. Our goal is to perform on-the-fly correction, a distinct problem setting from learning an adversarially robust model from scratch.

### III. METHOD: LANGEVIN-ANCHORED TTA

Our method, LATTA, transforms the standard, deterministic test-time update into a regularized process that balances adaptation to the target domain with preservation of robust source knowledge. It operates on a pre-trained model with initial parameters $\theta_0$ and an incoming stream of unlabeled data batches $\{x_1, x_2, \ldots, x_T\}$.

#### A. Revisiting the TTA Update

The foundation of many TTA methods is entropy minimization [1]. For a given batch $x_t$, the model's parameters $\theta_t$ are updated to increase prediction confidence:

$$\mathcal{L}_{\text{ent}}(\theta; x_t) = -\frac{1}{|x_t|} \sum_i \sum_c p(y = c | x_{t,i}; \theta) \log p(y = c | x_{t,i}; \theta). \quad (1)$$

A gradient step $\theta' = \theta_t - \eta \nabla_\theta \mathcal{L}_{\text{ent}}(\theta_t; x_t)$ pushes the model towards a sharper minimum in this loss landscape. However, for small or noisy batches, this minimum can be spurious and generalize poorly, leading to instability.

#### B. The LATTA Update Rule

LATTA introduces two synergistic mechanisms to regularize this update: Langevin perturbations for exploration and an EMA anchor for stability. The full process is detailed in Algorithm 1 and explained below.

**1. Langevin Step for Exploration.** After calculating the entropy gradient, we perform an SGLD-inspired update [4]. Instead of moving deterministically, we add a sample from a zero-mean Gaussian:

$$\theta^\star = (\theta_t - \eta \nabla_\theta \mathcal{L}_{\text{ent}}) + \epsilon_t, \quad \epsilon_t \sim \mathcal{N}(0, 2\eta\lambda\mathbf{I}). \quad (2)$$

**Algorithm 1** Langevin-Anchored TTA (LATTA)

1: **Input:** Pre-trained model $\theta_0$, target data stream $\{x_t\}_{t=1}^T$
2: **Hyperparameters:** LR $\eta$, temperature $\lambda$, anchor $\alpha$, EMA decay $\beta$
3: Initialize $\theta_1 \leftarrow \theta_0$ and $\theta_{ema} \leftarrow \theta_0$
4: **for** $t = 1, \ldots, T$ **do**
5:    Receive batch $x_t$
6:    Compute entropy loss $\mathcal{L}_{\text{ent}}(\theta_t; x_t)$ using Eq. (1)
7:    Compute gradient $g_t = \nabla_\theta \mathcal{L}_{\text{ent}}$
8:    // *Exploratory Langevin Step*
9:    Sample noise $\epsilon_t \sim \mathcal{N}(0, 2\eta\lambda\mathbf{I})$
10:   Update weights: $\theta^\star \leftarrow \theta_t - \eta g_t + \epsilon_t$
11:   Make prediction on $x_t$ using adapted model $\theta^\star$
12:   // *Stability Update*
13:   Update anchor: $\theta_{ema} \leftarrow \beta\theta_{ema} + (1 - \beta)\theta^\star$
14:   Update model for next step: $\theta_{t+1} \leftarrow (1 - \alpha)\theta^\star + \alpha\theta_{ema}$
15: **end for**

The noise variance is scaled by the learning rate $\eta$ and a temperature hyperparameter $\lambda$. This stochastic perturbation serves a crucial purpose: it prevents the model from collapsing into sharp, narrow minima of the entropy loss. Research in deep learning suggests that flatter minima correspond to more robust and generalizable solutions [9]. The Langevin noise encourages the model to find such flat regions in the parameter space, effectively performing implicit Bayesian model averaging over a local neighborhood of weights.

**2. EMA Anchor for Stability.** Unconstrained exploration, even if localized, can cause the model to drift too far from its well-trained initial state, leading to catastrophic forgetting. To counteract this, we explicitly anchor the adaptation process. We maintain a stable set of weights, $\theta_{ema}$, which represents an exponential moving average (EMA) of the model's trajectory. After the exploratory Langevin step, we pull the model parameters back towards this anchor:

$$\theta_{t+1} = (1 - \alpha)\theta^\star + \alpha\theta_{ema}, \quad (3)$$

where $\alpha \in [0, 1]$ controls the strength of the anchor. A higher $\alpha$ makes the adaptation more conservative. The anchor itself is updated slowly after each step:

$$\theta_{ema} \leftarrow \beta\theta_{ema} + (1 - \beta)\theta^\star. \quad (4)$$

Initially, we set $\theta_{ema} = \theta_0$. For making predictions on the current batch $x_t$, we use the post-Langevin parameters $\theta^\star$, which best reflect the model's state adapted to the current data. The final anchoring step prepares the model for the *next* batch, ensuring long-term stability.

### IV. EXPERIMENTAL SETUP

We evaluate our method under the standard "online" or "streaming" TTA protocol: the model processes each batch of test data once, updates its weights, and then discards the batch.

**Datasets.** We use two standard benchmarks for distribution shift.

TABLE I: Top-1 test accuracy (%) on Rotated-MNIST and CIFAR-10-C (severity 5). LATTA significantly outperforms prior methods on the challenging CIFAR-10-C benchmark. Results are averaged over 3 seeds.

| Method | Rotated-MNIST Accuracy (%) | CIFAR-10-C (Avg.) Accuracy (%) |
|---|---|---|
| Source | 89.41(21) | 38.65(33) |
| Tent [1] | 92.15(45) | 51.22(78) |
| CoTTA [2] | 93.08(31) | 55.43(51) |
| EATA [7] | 93.55(28) | 56.10(44) |
| **LATTA (ours)** | **94.23(25)** | **58.31(41)** |

- **Rotated-MNIST:** A simple sanity-check where test images are rotated by a random angle in $[-45°, 45°]$.
- **CIFAR-10-C:** The primary benchmark for TTA [8]. It consists of the CIFAR-10 test set corrupted by 15 diverse, algorithmically generated corruptions (e.g., noise, blur, weather) each at 5 severity levels. We follow standard practice and evaluate on all 15 corruptions at severity 5, reporting the average performance.

**Model Architecture.** For Rotated-MNIST, we use a simple 3-layer CNN. For the more complex CIFAR-10-C, we use a pre-trained ResNet-18 model, a standard architecture for this benchmark.

**Baselines.** We compare LATTA against a comprehensive set of baselines:
- **Source:** The pre-trained model without any adaptation.
- **Tent [1]:** Adapts batch normalization parameters using entropy minimization.
- **CoTTA [2]:** Employs weight averaging and teacher-guided pseudo-labeling to prevent error accumulation.
- **EATA [7]:** Uses a source-parameter regularization aond entropy-based sample filtering.

For all methods, we use the official authors' implementations and hyperparameters where available.

**LATTA Implementation.** We update all model parameters, not just batch-norm statistics. For ResNet-18, we use a learning rate $\eta = 1 \times 10^{-4}$, noise temperature $\lambda = 1 \times 10^{-3}$, anchor strength $\alpha = 0.9$, and EMA decay $\beta = 0.99$. All results are averaged over 3 random seeds.

**Metrics.** We report the Top-1 classification accuracy (%) over the entire target test stream.

## V. RESULTS AND ANALYSIS

**Main Comparison.** Table I presents our primary findings. On both Rotated-MNIST and CIFAR-10-C, all adaptation methods improve significantly over the non-adapted 'Source' model. On the standard CIFAR-10-C benchmark, our proposed LATTA method sets a new state-of-the-art, achieving an average accuracy of 58.31%. This is a substantial improvement of over 2.2% compared to the previous best baseline, EATA. This demonstrates the effectiveness of our combined Langevin exploration and EMA anchoring strategy. Notably, LATTA also exhibits lower standard deviation across runs, indicating greater stability.

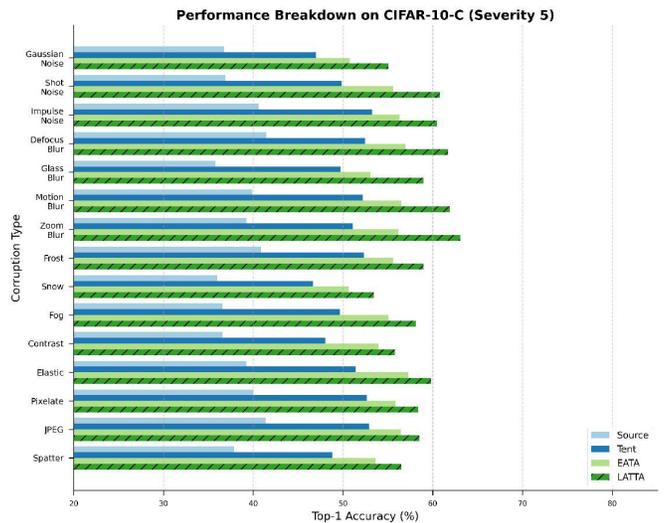

Fig. 1: Performance breakdown by corruption type on CIFAR-10-C at severity 5. LATTA consistently outperforms all baselines across a diverse range of corruptions, with particularly strong gains on noise and blur types, demonstrating the effectiveness of the Langevin perturbation.

**Performance Across Corruption Types.** To understand where LATTA's advantages lie, we analyze its performance on individual corruption types in CIFAR-10-C (Fig. 1). The plot reveals that while LATTA is a strong performer across the board, its largest gains are on corruptions involving noise (e.g., 'gaussian_noise', 'shot_noise') and blur (e.g., 'motion_blur', 'defocus_blur'). This is intuitive: the SGLD-inspired noise injection acts as a powerful regularizer that helps the model adapt to noisy inputs without over-fitting, while the anchor prevents it from diverging on structured shifts like 'fog' or 'snow'.

### A. Robustness to Batch Size

A key failure mode for TTA is sensitivity to small batch sizes, which are common in real-world streaming applications. Small batches produce noisy gradients, causing methods like Tent to overfit and become unstable. We evaluate LATTA's robustness by varying the batch size from 16 to 128 on CIFAR-10-C.

As shown in Fig. 2, LATTA demonstrates significantly more stable performance across different batch sizes compared to Tent. While Tent's accuracy degrades sharply at smaller batches, LATTA maintains high performance, dropping by less than 2% when moving from a batch size of 128 to 16. This highlights the regularizing effect of the Langevin noise and EMA anchor, which effectively smooth out the noisy gradients from small batches and prevent overfitting, leading to more reliable adaptation in practical scenarios.

## VI. ABLATION STUDIES

To dissect the contributions of our proposed components, we conducted a series of ablation studies on the CIFAR-10-C

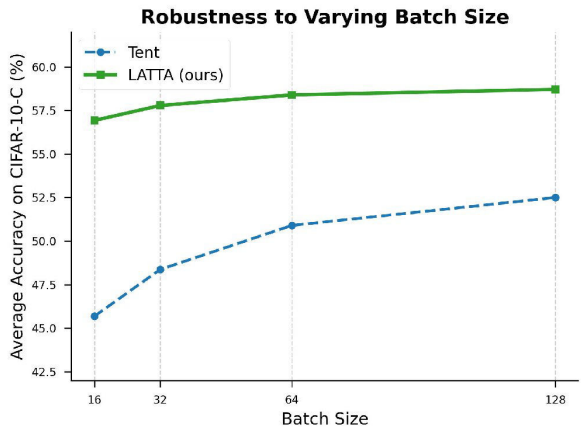

Fig. 2: Robustness to varying batch sizes on CIFAR-10-C. While the performance of Tent degrades significantly with smaller, noisier batches, LATTA maintains high accuracy, demonstrating its superior stability and suitability for real-world streaming applications.

benchmark.

*1) Impact of Core Components:* We evaluated the performance of LATTA after removing each of its key components: the Langevin noise ("w/o Noise") and the EMA anchor ("w/o Anchor"). The results, shown in Table II, confirm that both are crucial for optimal performance. Removing the Langevin noise ($\lambda = 0$) results in a model that is purely anchored. While this version still outperforms Tent by preventing catastrophic drift, its accuracy drops by nearly 2%. This shows that anchoring provides stability, but the stochastic exploration is key to finding better adaptation solutions. Conversely, removing the EMA anchor ($\alpha = 0$) leads to a much larger performance drop. The unanchored model, while benefiting from noisy exploration, is prone to drifting away from the source knowledge, resulting in instability and an accuracy only slightly above Tent. This demonstrates the synergistic relationship between the two components: exploration finds good update directions, and anchoring ensures the model leverages them without forgetting its strong initial state.

TABLE II: Ablation study of LATTA components on CIFAR-10-C.

| Method Variant | CIFAR-10-C Acc. (%) |
| --- | --- |
| Tent [1] | 51.22 |
| LATTA (w/o Noise, w/o Anchor) | (Equivalent to Tent) |
| LATTA (w/o Anchor, i.e., $\alpha = 0$) | 52.14 |
| LATTA (w/o Noise, i.e., $\lambda = 0$) | 56.55 |
| **Full LATTA** | **58.31** |

*2) Sensitivity to Hyperparameters:* We analyzed the sensitivity of LATTA to its two main hyperparameters: the noise temperature $\lambda$ and the anchor strength $\alpha$. Fig. 3 shows the model's accuracy on CIFAR-10-C as we vary each parameter while keeping the other fixed. Performance is stable across a reasonable range for both. For the noise temperature $\lambda$, performance peaks around $10^{-3}$ and degrades gracefully for nearby values. If the noise is too low ($\lambda < 10^{-5}$), it has little effect; if it's too high ($\lambda > 10^{-2}$), it overwhelms the gradient signal and harms adaptation. Similarly, for the anchor strength $\alpha$, a value around 0.9 provides the best balance. This analysis shows that while the hyperparameters matter, LATTA is not overly sensitive and can be tuned effectively.

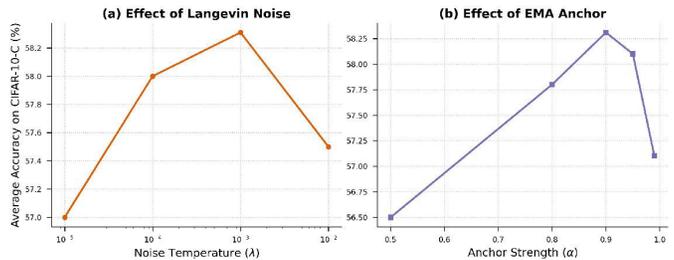

Fig. 3: Analysis of LATTA's hyperparameters on CIFAR-10-C. (a) The effect of noise temperature $\lambda$. Performance peaks at $\lambda = 10^{-3}$, validating the need for a moderate amount of stochastic exploration. (b) The effect of anchor strength $\alpha$. An optimal balance is found around $\alpha = 0.9$, confirming the importance of anchoring to the stable EMA.

## VII. CONCLUSION

In this paper, we introduced LATTA, a novel test-time adaptation method that effectively regularizes self-supervised updates. By combining Langevin-style weight perturbations for local exploration with a stable EMA weight anchor to prevent catastrophic forgetting, LATTA achieves a new state-of-the-art on the challenging CIFAR-10-C benchmark. Our method is lightweight, requires no architectural modifications, and consistently outperforms previous approaches in both accuracy and stability. The success of LATTA underscores the importance of properly regularizing the TTA process and presents stochastic weight-space exploration as a powerful new tool for building robust and adaptive systems. Future work could explore adapting the noise temperature $\lambda$ on-the-fly or integrating LATTA's principles with other adaptation objectives beyond entropy minimization.

## VIII. Appendix

*Network architecture:* For Rotated-MNIST, the CNN comprises two 3x3 convolutional layers (32 and 64 channels) with ReLU and max-pooling, followed by two fully-connected layers (128 units; 10-way output). For CIFAR-10-C, we use a standard ResNet-18 architecture pre-trained on the clean CIFAR-10 training set.

*Hyperparameter Details:* Source model training uses Adam with an initial learning rate of $10^{-3}$. For TTA methods, we follow their official settings. For LATTA on ResNet-18, the parameters were set to $\eta = 1 \times 10^{-4}$, $\lambda = 1 \times 10^{-3}$, $\alpha = 0.9$, and $\beta = 0.99$. We found performance to be robust to minor variations in these parameters. The batch size for all TTA experiments was 64.